\documentclass[10pt,twocolumn,letterpaper]{article}

\usepackage[margin=0.9in]{geometry}
\usepackage{times}
\usepackage{epsfig}
\usepackage{graphicx}
\usepackage{amsmath}
\usepackage{amssymb}

\usepackage{subcaption}
\usepackage{amsmath}
\usepackage{amsfonts}
\usepackage{booktabs} 
\usepackage{float}

\usepackage{sectsty}

\sectionfont{\fontsize{13}{15}\selectfont}

\usepackage[pagebackref=true,breaklinks=true,letterpaper=true,colorlinks,bookmarks=false]{hyperref}

\graphicspath{{figures/}}

\begin{document}

\title{Deep Learning with Inaccurate Training Data for Image Restoration}

\author{Bolin Liu\\
McMaster University\\
{\tt\small liub30@mcmaster.ca}
\and
Xiao Shu\\
Shanghai Jiao Tong University\\
{\tt\small shux@sjtu.edu.cn}
\and
Xiaolin Wu\\
Shanghai Jiao Tong University\\
{\tt\small xwu510@sjtu.edu.cn}
}

\date{}

\maketitle
\thispagestyle{empty}

\begin{abstract}
  In many applications of deep learning, particularly those in image
  restoration, it is either very difficult, prohibitively expensive,
  or outright impossible to obtain paired training data precisely as
  in the real world.  In such cases, one is forced to use synthesized
  paired data to train the deep convolutional neural network (DCNN).
  However, due to the unavoidable generalization error in statistical
  learning, the synthetically trained DCNN often performs poorly on
  real world data.  To overcome this problem, we propose a new general
  training method that can compensate for, to a large extent, the
  generalization errors of synthetically trained DCNNs.
\end{abstract}

\section{Introduction}
\label{sec:intro}

Over the past few years deep learning has become a widely advocated
approach for image restoration.  A large number of research papers
have been published on the use of deep convolutional neural networks
(DCNN) for image super-resolution, denoising, deblurring,
demosaicking, descreening, etc.  All these authors reported great
improvements of the restoration results by the DCNN methods over
traditional image processing methods.  However, as demonstrated by
this work, there is still a technical hurdle to be cleared, before one
can firmly establish the superiority of data-driven deep learning
approach to the traditional model-based inverse problem framework for
image restoration.

\begin{figure}[t]
  \captionsetup[subfigure]{labelformat=empty}
  \centering
  \begin{subfigure}{0.3\linewidth}
    \centering
    \includegraphics[width=\linewidth]{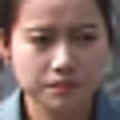}
    \caption{Bicubic}
  \end{subfigure}
  \begin{subfigure}{0.3\linewidth}
    \centering
    \includegraphics[width=\linewidth]{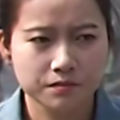}
    \caption{EDSR}
  \end{subfigure}
  \begin{subfigure}{0.3\linewidth}
    \centering
    \includegraphics[width=\linewidth]{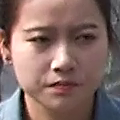}
    \caption{Ground truth}
  \end{subfigure}
  \begin{subfigure}{0.3\linewidth}
    \centering
    \includegraphics[width=\linewidth]{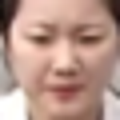}
    \caption{Bicubic}
  \end{subfigure}
  \begin{subfigure}{0.3\linewidth}
    \centering
    \includegraphics[width=\linewidth]{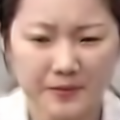}
    \caption{EDSR}
  \end{subfigure}
  \begin{subfigure}{0.3\linewidth}
    \centering
    \includegraphics[width=\linewidth]{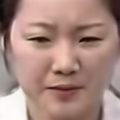}
    \caption{\textbf{EDSR+}}
  \end{subfigure}
  \caption{Sample $\times 4$ super-resolution results of the original
    EDSR model and our refined version, EDSR+.  Although EDSR performs
    well for synthetic low-resolution input image (first row), it is
    ineffective for real image (second row) in comparison with EDSR+}
  \label{fig:sr_sample}
\end{figure}

In the vast majority of the published studies on deep learning image
restoration, the pairs of high-quality and degraded images for
training the neural networks are generated synthetically using a
simplistic degradation model.  For example, in all deep learning
super-resolution papers, the low-resolution images are generated from
their full resolution counterparts via some downsampling kernel like
bicubic.  But the actual physical formations of such images are far
more complex and compounded, involving various factors such as the
camera point spread function (PSF), pixel sensors crosstalk,
demosaicking, compression, camera/object motion, etc.  The discrepancy
between the downsampling operator and the true signal degradation
function is more than enough to derail these learning methods.  As
exemplified in Figure~\ref{fig:sr_sample}, although the existing deep
learning super-resolution methods perform very well on artificially
generated low-resolution images, they cannot super-resolve real-world
images nearly at the same quality.  Unfortunately, many outstanding
image restoration results reported in the literature, which are
obtained with perfect match of the data for training and inference in
statistics, are somewhat misleading and irreproducible on real images.

A more sophisticated data synthesizer is certainly warranted to
realize the full potential of a deep learning method on real data.
But in many applications of image restoration, it is very difficult,
if not impossible, to simulate the complex compound effects of
multiple degradation causes, some of which may be stochastic, to the
desired precision.  To further demonstrate this difficulty in the
paper, we also study the problem of demoir\'e, the task of removing
the interference patterns in a camera-captured screen image, in
addition to super-resolution.  In this case, given an original image,
its degraded version after being displayed on screen and then captured
by camera may be synthesized approximately for training using a very
complex image formation model, but the accuracy of the synthesis is
still insufficient due to intricate interplays of lens distortion,
defocus blur, screen glass glare, etc.

Very recently, the issue of statistically mismatched training data in
DCNN image denoising methods was getting brought up and discussed
\cite{plotz2017benchmarking, anaya2018renoir, chen2018image,
  xu2018real}.  After pointing out the pitfalls of using simple
Gaussian noise model to create noisy and clean image pairs to train
DCNNs, these authors proposed to alleviate the problem by generating
more realistic paired training images, either using data acquired in
extensive experiments with an assortment of cameras operating in
different ISO settings, or using generative adversarial network (GAN)
to combine clean images with noise extracted from real noisy images.
However, the paired training data generated by the above improved
techniques are still synthesized after all; inevitably, they deviate
from those in the real world situations.


As outlined above, the most likely scenario for a deep learning image
restoration system is that, even with the best degradation model $M$
on hand, artificially degraded image $Y_s = M(X)$ from clean image $X$
still differs greatly from real degraded image $Y$.  Thus, although
the synthetic-real image pairs $(Y_s;X)$ can be used to train a
generative DCNN $G_0$ directly as in current practice, the resulting
$G_0$ is often ineffective for real images.

In our design, the algorithm development does not end with $G_0$.
Instead, the synthetically trained restoration DCNN $G_0$ is used to
produce so-called surrogate ground truth images $X_s$ to pair up with
real degraded input images $Y$, when the real ground truth images $X$
for $Y$ cannot be obtained.  In other words, we feed $Y$ into the DCNN
$G_0$ to generate the corresponding restored images $X_s = G_0(Y)$.
The real degraded images $Y$ and their surrogate ground truth images
$X_s$, the latter being approximations of $X$, provide the paired data
$(Y;X_s)$ to refine the restoration network $G_0$ through supervised
learning.

We would like to stress a vital difference between the existing
methods and our method.  Previous authors used real pristine images
$X$ but paired them with synthetic degraded images in the training of
restoration DCNNs.  In contrast, we stick to real degraded images $Y$
and pair them with synthetic ground truth images $X_s$.  We argue,
also as demonstrate by empirical results, that the latter approach is
the correct one.

When practical limitations prevent the deep learning algorithm
designers from having the ideal paired training data $(Y;X)$ that are
exact as the physical reality, they have two options as to choosing
paired training data: $(Y_s;X)$ or $(Y;X_s)$.  There is an obvious
paradox in using the paired synthetic-real data $(Y_s;X)$ to train the
restoration DCNN.  At the time of inference, the DCNN needs to
estimate the latent image $X$ from the real degraded image $Y$, not
from the synthetic one $Y_s$ as in the training.  Moreover, it is
operationally very difficult, if not impossible, in the DCNN design
stage to account for the imprecisions in the synthetic input data
$Y_s$.

On the other hand, by taking the second option, we train the
restoration DCNN to work with the real input degraded images $Y$.
Although in the training stage, we have to use surrogate ground truth
images $X_s$ at the output end of the network, we can easily include
suitable penalty terms in the objective function to remove or
alleviate the effects of differences between $X_s$ and $X$.  An
immediate idea is to use the GAN technique to augment the signal level
fidelity of the paired data by the statistics of unpaired clean
images.

In addition to the above highlighted contribution, the proposed new
training method is a general one; it can be applied to improve any
existing DCNNs for image restoration, regardless the specific
application problems addressed.

\section{Related Work}
\label{sec:related}

The problem of statistical differences between synthetic and real data
has long been overlooked in the literature of learning based image
restoration.  Not until recently have there been a few attempts to
alleviate the problem in some specific applications, such as denoising
and super-resolution.  Most of these studies focus on improving the
truthfulness of their training data synthesizers.  For instance, to
generate artificial noise for multi-image denoising,
\cite{mildenhall2018burst} first converts images to linear color space
with real data calibrated invert gamma correction and then generates
noise with distribution estimated from real images.  For single image
denoising, \cite{guo2018toward} employs a sophisticated noise
synthesizer that takes multiple factors into consideration, such as
signal-dependent noise and camera processing pipeline.  In
\cite{chen2018image}, the authors proposed a GAN-based neural network
to generate realistic camera noise.  The network is trained with
high-quality images superimposed with noise patterns extracted from
smooth regions of real noisy images, assuming that the high frequency
components of these smooth regions only come from sensor noise and
noise is independent to signal.  These assumptions however are
impractical and restrictive, making it difficult to extend the idea
for other image restoration problems.

The problem of inaccurate degradation model is also recognized by the
authors of \cite{shocher2017zero} in their study of data-driven
super-resolution.  To alleviate the problem, they proposed a
relatively shallow neural network, called ZSSR, which is trained only
with patches aggregated from the input image.
However, ZSSR still relies on bicubic downsampling to synthesize the
corresponding low-resolution patches; the problem of the unrealistic
downsampler is left unaddressed.

Another possible option is to give up explicit degradation model
completely and use unpaired training instead.  CycleGAN provides a
mechanism to construct a bi-directional mapping between two types of
images without paired samples \cite{zhu2017unpaired, yi2017dualgan}.
However, the learning of the degradation process can be extremely
challenging without precisely aligned data let alone restoration.
Thus, it is difficult to achieve highly accurate results using this
approach.

Due to the complexity of image formation, it is extremely difficult to
design a good data synthesizer in many image restoration applications.
Even the seemingly minor details of a synthetic process, such as
whether the noise values are rounded to integers, can have a
significant impact on the performance of a deep learning technique
\cite{chen2015learning}.  Thus, instead of using data synthesis, many
studies try to use extra images captured in non-ideal conditions as
the input.  For example, by using different camera exposure settings,
one can obtain a series of images of the same scene with varied noise
levels and then use the underexposed ones as the noisy input
\cite{plotz2017benchmarking}.  To train a deblurring DCNN, one can
average all the frames of a high frame-rate video clip as the blurry
input and pick one of the frames as the sharp ground truth image
\cite{nah2017deep}.  While employing extra images mitigates the
problem of inaccurate synthesizer, it suffers from several serious
drawbacks.  First, due to the nature of using multiple images, the
imaged subject must stay perfectly still, and complex image processing
operators, such as lens correction, spatial alignment, intensity
matching are required to couple the high-quality and degraded image.
Second, as it is often prohibitively expensive to build a sufficiently
large data set by taking images one by one, a network can overfit the
available data and become sensitive to particular camera brands,
exposure settings and captured scenes \cite{xu2018real}.

\section{Two-Stage Training}
\label{sec:algorithm}

\begin{figure}
  \centering
  \includegraphics[width=0.98\linewidth]{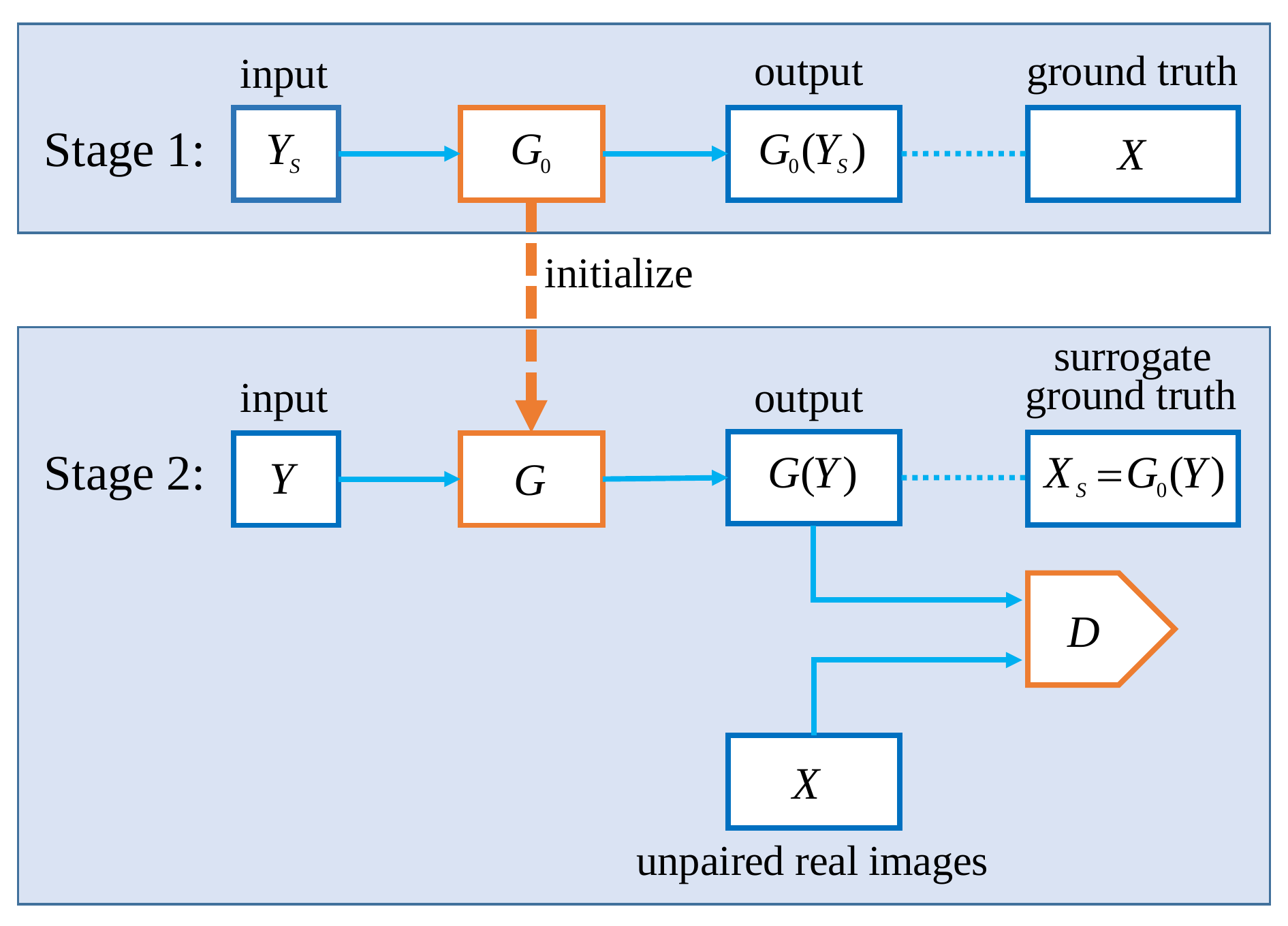}
  \caption{The proposed two-stage learning approach.}
  \label{fig:algo_sketch}
\end{figure}

Due to the lack of paired real data, most existing deep learning image
restoration techniques train their DCNNs $G_0$ with artificially
degraded images $Y_s$ paired with the corresponding high-quality
ground truth images $X$.  However, as we discussed previously, this
mainstream approach is fundamentally flawed: a DCNN, which has never
``seen'' any real degraded images in training, performs
unsatisfactorily for real images.  In this section, we introduce an
alternative training approach, which circumvents the aforementioned
problem by using real degraded images $Y$ and surrogate ground truth
images $X_s$ as the training data.

In most applications, collecting a large number of real degraded
images $Y$ that are governed by the same distribution as the real
input is not difficult, but it is a challenging task to find an
accurate estimate $X_s$ of the corresponding latent ground truth image
$X$.  After all, that is exactly the restoration problem that we want
to solve.  However, even if we have to resort to inaccurate surrogate
ground truth images $X_s$ in the training of a DCNN $G$, there are
still various ways to rectify the final output $G(Y)$.  In other
words, with fitting image priors and properly designed penalty terms,
imperfect the surrogate ground truth images might not severely affect
the training of $G$ for real degraded input images.  Thus, we can
simply train a DCNN $G_0$ as the synthesizer using artificially
degraded images $Y_s$ and real ground truth $X$, and then generate the
surrogate ground truth images $X_s$ by feeding $G_0$ with real
degrades images $Y$, i.e., let $X_s = G_0(Y)$.

Based on the above idea, we design a two-stage training approach, as
sketched in Figure~\ref{fig:algo_sketch}.  In the first stage, the
DCNN $G_0$ is trained using the conventional approach with synthetic
input $Y_s$ and real ground truth $X$.  Then in the second stage, the
network is retrained as $G$ using real degraded image $Y$ and
surrogate ground truth $G_0(Y)$ generated by the first stage network
$G_0$.  Due to the similarities of their training data, the second
training stage can be significantly expedited by initializing $G$
using the weights of $G_0$.  While the synthetically trained $G_0$ can
produce a rough estimate $X_s$ of the latent ground truth for any
given real degraded image $Y$, the output $G_0(Y)$ is often plagued
with artifacts.  To prevent these objectionable artifacts from being
picked up by $G$, we include additional penalty terms, such as GAN, in
the objective function of the second stage.

The proposed training approach is a general one, applicable to any
existing image restoration DCNN techniques regardless of their network
architectures or objective functions.  For instance, many restoration
techniques optimize pixel-wise loss function, such as mean squared
error (MSE), in the training of their networks, as follows,
\begin{align}
  L_2 &= \frac{1}{WH}\|G_0(Y_s) - X\|_2^2,
\end{align}
where the $W$ and $H$ are the width and height of the network output,
respectively.  With our two-stage training approach, we can use these
objective functions unaltered in the first training stage.  Since the
network architecture also requires no modification, the trained model
from the original authors, if available, can be used as $G_0$ directly
without retraining the network.  In the second stage, the training set
is consist of real-synthetic pairs $(Y; X_s)$ instead of
synthetic-real pairs $(Y_s; X)$.  Correspondingly, the example MSE
loss function can be written as,
\begin{align}
  L_2 \!=\! \frac{1}{W \! H}\|G(Y) \!-\! X_s\|_2^2
  \!=\! \frac{1}{W \! H}\|G(Y) \!-\! G_0(Y)\|_2^2.
\end{align}


In this second stage, the fidelity terms in objective function
penalizes the differences between the output $G(Y)$ and the surrogate
ground truth $X_s$, pushing $G$ to produce similar results as network
$G_0$.  As network $G_0$ is essentially a trained inverse of the
degradation synthesizer function $M$, these fidelity terms implicitly
regulates $G$ as an inverse of $M$ as well.

In addition to the fidelity terms, we also need to ensure the output
$G(Y)$ is similar to real ground truth images statistically.  Without
the matched ground truth $X$ to $Y$, a possible solution to the
problem is to use unpaired learning techniques.  In the proposed
training approach, we integrate the GAN technique in the second stage
\cite{goodfellow2014generative}.  In GAN, a discriminative network $D$
is jointly trained with the generative network $G$ for discriminating
the output images $G(Y)$ against a set of unpaired clean images $X$.
This process guides $G$ to return images that are statistically
similar to artifact-free ground truth images without using paired
training data.

Following the idea of GAN, we set the discriminative network $D$ to
solve the following minimax problem:
\begin{align}
  \min_G \max_D (
  & \mathbb{E}_X \left[ \log D(X) \right]
\nonumber
\\
& + \mathbb{E}_Y \left[ \log(1-D(G(Y))) \right]).
\end{align}
This introduces an adversarial term in the loss function of the
generator network $G$:
\begin{align}
  L_{A} = -\log D(G(Y)).
\end{align}
In competition against generator network $G$, the loss function for
training discriminative network $D$ is the binary cross entropy:
\begin{align}
  L_D = - \left[ \log(D(X))+\log(1-D(G(Y))) \right].
\end{align}
Minimizing $L_{A}$ drives network $G$ to produce images that
network $D$ cannot distinguish from original artifact-free images.
Accompanying the evolution of $G$, minimizing $L_D$ increases the
discrimination power of network $D$.

Combining the MSE loss $L_2$ and adversarial loss $L_{A}$, we
arrive at the final formulation of the loss function for optimizing
$G$ in the second training stage,
\begin{equation}
  L_G = L_2 + \lambda L_{A},
\label{eq:joint training}
\end{equation}
where Lagrange multiplier $\lambda$ is user given weight balancing the
two terms.

In the next two sections, we apply the proposed two-stage training
approach in two very different image restoration problems to showcase
the efficacy and generality of the idea.



\section{Experiment on Face Super-Resolution}
\label{sec:facesr}

Super-resolution is one of the most intensively researched image
restoration problems.  In the last few years, DCNN-based
Super-resolution techniques have demonstrated its great potential and
significantly advanced the state of the art.  However, almost all of
these techniques still rely on unrealistic degradation models, such as
bicubic downsampling, to synthesize training data, causing
generalization problems in real-world applications.  In this section,
we select two popular super-resolution techniques, SRResNet
\cite{ledig2016photo} and EDSR \cite{lim2017enhanced}, and try to
improve their performance on real low-resolution images with the
proposed two-stage training approach.  We also test ZSSR
\cite{shocher2017zero}, a super-resolution technique designed
specifically for real images, and SRGAN \cite{ledig2016photo}, a
GAN-based super-resolution technique, as references.

\subsection{Preparation of Training Data}

There are plenty of human face data sets available to researchers, but
images in most of these sets were captured in controlled laboratory
environment and none of them provides the required real low-resolution
images for this research.  Thus, we have to build our own face image
data sets.  The face images in our data set are cropped directly from
video clips; no other image processing operators is used.  The imaged
persons are ethnically Asian, in various age groups and with different
facial expressions and poses.  The extracted face images are in two
different resolutions: high-resolution ($120 \times 120$) and
low-resolution ($30 \times 30$).  The high-resolution images are used
as the real ground truth images $X$, and the low-resolution images are
used as the real degraded images $Y$.  The artificially degraded
images $Y_s$ are scaled down from $X$ using $\times 4$ bicubic
downsampling.

\subsection{Training Details}


Of the 60,000 high-resolution face images in our data set, 40,000 are
used for training SRResNet, EDSR and SRGAN; 10,000 images are for
validation; and another 10,000 are for testing.  The improved versions
of SRResNet and EDSR by our training approach are called SRResNet+ and
EDSR+ respectively.  In the second training stage, SRResNet+ and EDSR+
use 10,000 real low-resolution images as input and 10,000
high-resolution images as unpaired real images for GAN.  Moreover,
SRGAN is also trained with the same unpaired real images for its GAN.

All the tested techniques, SRResNet, EDSR, ZSSR and SRGAN, are trained
using the recommended settings by their authors.  The loss function of
SRResNet+ and EDSR+ in the second training stage is set as,
\begin{equation}
  L_G = L_2
  + 10^{-3} \times L_{A},
\label{eq:G_loss}
\end{equation}
and the learning rate and batch size are set as $10^{-4}$ and 32,
respectively.  The second training stage takes $10^4$ update
iterations, which are $1/10$ of the amount used by the first stage.
Thus, the required time for the second training stage is roughly
$1/10$ of the first.  For GAN training, we set $k=1$ and update $G$
and the discriminator $D$ alternatively.

\subsection{Experimental Results}

\begin{figure}
  \captionsetup[subfigure]{labelformat=empty}
  \centering
  \begin{subfigure}{0.24\linewidth}
    \centering
    \includegraphics[width=\linewidth]{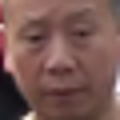} \\
    \includegraphics[width=\textwidth]{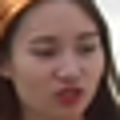}
    \caption{Bicubic}
  \end{subfigure}
  \begin{subfigure}{0.24\linewidth}
    \centering
    \includegraphics[width=\linewidth]{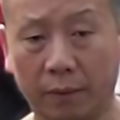} \\
    \includegraphics[width=\textwidth]{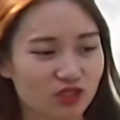}
    \caption{SRResNet}
  \end{subfigure}
  \begin{subfigure}{0.24\linewidth}
    \centering
    \includegraphics[width=\linewidth]{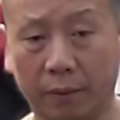} \\
    \includegraphics[width=\textwidth]{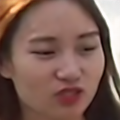}
    \caption{EDSR}
  \end{subfigure}
  \begin{subfigure}{0.24\linewidth}
    \centering
    \includegraphics[width=\linewidth]{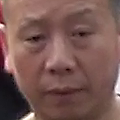} \\
    \includegraphics[width=\textwidth]{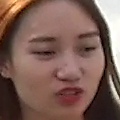}
    \caption{Ground truth}
  \end{subfigure} \\
  \caption{Results of $\times 4$ super-resolution for synthetic
    low-resolution images.}
  \label{fig:syn_sr}
\end{figure}

As shown in Figure ~\ref{fig:syn_sr}, both SRResNet and EDSR perform
well on our synthetic data set, scoring average peak signal-to-noise
ratio (PSNR) 30.98 dB and 31.10 dB, respectively.  Their output images
are very similar to the ground truth images visually.

\begin{figure*}[h]
  \captionsetup[subfigure]{labelformat=empty}
  \centering
  \begin{subfigure}{0.133\linewidth}
    \centering
    \includegraphics[width=\linewidth]{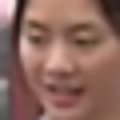} \\
    \includegraphics[width=\linewidth]{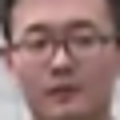} \\
    \includegraphics[width=\linewidth]{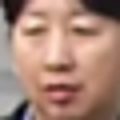} \\
    \includegraphics[width=\linewidth]{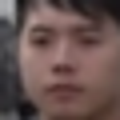} \\
    \includegraphics[width=\linewidth]{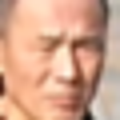}
    \caption{Bicubic}
  \end{subfigure}
  \begin{subfigure}{0.133\linewidth}
    \centering
    \includegraphics[width=\linewidth]{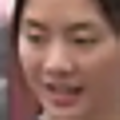} \\
    \includegraphics[width=\linewidth]{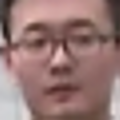} \\
    \includegraphics[width=\linewidth]{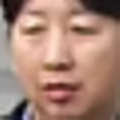} \\
    \includegraphics[width=\linewidth]{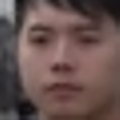} \\
    \includegraphics[width=\linewidth]{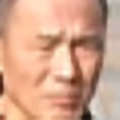}
    \caption{ZSSR}
  \end{subfigure}
  \begin{subfigure}{0.133\linewidth}
    \centering
    \includegraphics[width=\linewidth]{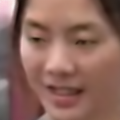} \\
    \includegraphics[width=\linewidth]{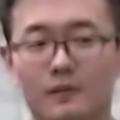} \\
    \includegraphics[width=\linewidth]{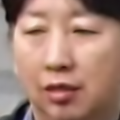} \\
    \includegraphics[width=\linewidth]{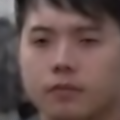} \\
    \includegraphics[width=\linewidth]{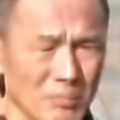}
    \caption{SRResNet}
  \end{subfigure}
  \begin{subfigure}{0.133\linewidth}
    \centering
    \includegraphics[width=\linewidth]{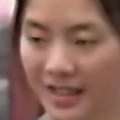} \\
    \includegraphics[width=\linewidth]{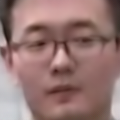} \\
    \includegraphics[width=\linewidth]{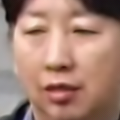} \\
    \includegraphics[width=\linewidth]{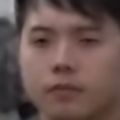} \\
    \includegraphics[width=\linewidth]{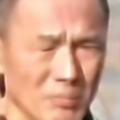}
    \caption{SRGAN}
  \end{subfigure}
  \begin{subfigure}{0.133\linewidth}
    \centering
    \includegraphics[width=\linewidth]{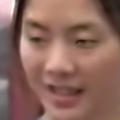} \\
    \includegraphics[width=\linewidth]{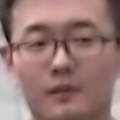} \\
    \includegraphics[width=\linewidth]{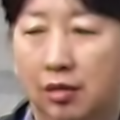} \\
    \includegraphics[width=\linewidth]{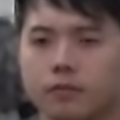} \\
    \includegraphics[width=\linewidth]{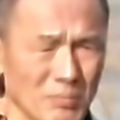}
    \caption{EDSR}
  \end{subfigure}
  \begin{subfigure}{0.133\linewidth}
    \centering
    \includegraphics[width=\linewidth]{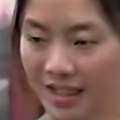} \\
    \includegraphics[width=\linewidth]{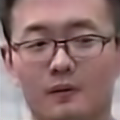} \\
    \includegraphics[width=\linewidth]{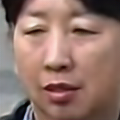} \\
    \includegraphics[width=\linewidth]{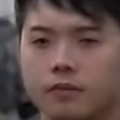} \\
    \includegraphics[width=\linewidth]{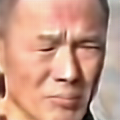}
    \caption{\textbf{SRResNet+}}
  \end{subfigure}
  \begin{subfigure}{0.133\linewidth}
    \centering
    \includegraphics[width=\linewidth]{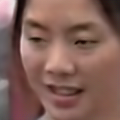} \\
    \includegraphics[width=\linewidth]{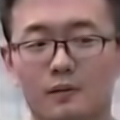} \\
    \includegraphics[width=\linewidth]{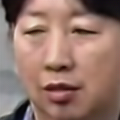} \\
    \includegraphics[width=\linewidth]{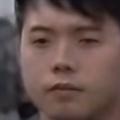} \\
    \includegraphics[width=\linewidth]{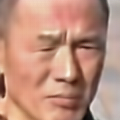}
    \caption{\textbf{EDSR+}}
  \end{subfigure} \\
  \vspace{0.2ex}
  \caption{Results of $\times 4$ super-resolution for real
    low-resolution images.}
  \label{fig:exp_sr}
\end{figure*}

However, as shown in Figure~\ref{fig:exp_sr}, neither SRResNet nor
EDSR can super-resolve real low-resolution images to the same quality
level as the synthetic one.  Even SRGAN, the GAN optimized version of
SRResNet, is incapable of improving the performance for real images.
Although to a human viewer, there are hardly any differences between
the real and synthetic low-resolution images, the super-resolution
results by the conventionally trained techniques are substantially
different.  In comparison, SRResNet+ and EDSR+, which are trained
using the proposed approach, have no difficulty in real image
super-resolution.  Their results recover lots of details and look
similar to the results of SRResNet and EDSR for synthetic images.
Please note that the improved techniques share the exact same network
architectures as the original versions; only weights are adjusted for
dealing with real images.

We also test ZSSR, which only needs the low-resolution input image for
training and testing.  To make a fair comparison, we use the whole
frame, instead of only a face image, as the input.  However, its
results are still inferior than the results of the other tested
techniques.

\section{Experiment on Demoir\'eing}
\label{sec:demoire}

\begin{figure*}[h!]
  \centering
  \includegraphics[width=\linewidth]{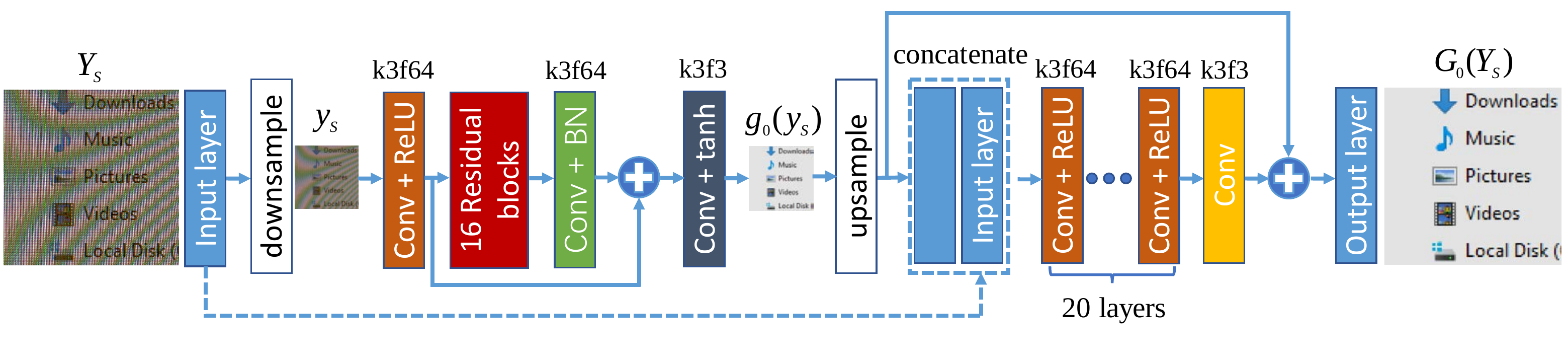}
  \caption{The architecture of the proposed DM-Net.  The
    kernel size $k$ and the number of feature maps $f$ are marked
    above each convolutional layer.}
  \label{fig:dmnet}
\end{figure*}

In this section, we demonstrate the proposed two-stage training
technique for another image restoration problem: demoir\'eing for
camera-captured screen images.  Taking photos of optoelectronic
displays is a direct and spontaneous way of transferring data and
keeping records, which is widely practiced.  However, due to the
analog signal interference between the pixel grids of the display
screen and camera sensor array, objectionable moir\'e (alias) patterns
appear in captured screen images.  As the moir\'e patterns are
structured, highly variant and correlated with signal, they are
difficult to be completely removed without affecting the underneath
latent image.  Despite the commonness and annoyance of the problem,
little work has been done on the reduction of moir\'e artifacts in
camera-captured screen images.

To solve this challenging problem, we purposefully designed a
multi-scale DCNN, called DM-Net.  As shown in Figure~\ref{fig:dmnet},
an input moir\'e image is first downsampled by a scale factor of 4 in
DM-Net and demoir\'ed by 16 deep residual blocks as in
\cite{gross2016training}.  The result is then upsampled to the
original resolution and further cleaned up by 20 more convolutional
layers.  We also experimented with two excellent general purpose image
restoration networks, DnCNN \cite{zhang2017beyond} and RED-Net
\cite{mao2016image}, and retrained them for the demoir\'eing task
using synthetic moir\'e images.  While these techniques can produce
more-or-less acceptable results for synthetic moir\'e images, their
performances deteriorate significantly when dealing with real images.

We tried to solve the problem for real images using CycleGAN
\cite{zhu2017unpaired, yi2017dualgan}.  It is however problematic for
this task.  As CycleGAN maps image $x$ to its clean version $G_{C}(x)$
then back ($F(G_{C}(x))=x$), it requires a one-to-one correspondence
between the two images.  But this does not hold for moir\'e images;
there can be multiple images of the same content but vastly different
moir\'e patterns.  Moreover, without using paired data, the inverse
mapping $F$ is unable to learn the degradation process precisely.  As
a result, the outputs of $F$ never become as realistic as those of our
carefully designed moir\'e pattern synthesizer.  Therefore, CycleGAN
does not make an improvement over the existing results.

In the section, we demonstrate the efficacy of the proposed two-stage
training approach for real image demoir\'ing.



\subsection{Preparation of Training Data}
\label{sec:data}


Ideally, the training process should only use real photographs of a
screen and the corresponding original digital images displayed on it.
While obtaining such a pair of images is easy, perfectly aligning them
spatially, a necessary condition for preventing mismatched edges being
misidentified as moir\'e patterns, is difficult to achieve.  As many
common imaging problems in real photos, such as lens distortion and
non-uniform camera shake, adversely affect the accuracy of image
alignment, it is challenging to build a sufficiently large and high
quality training set using real photos.

\begin{figure}[t]
  \centering
  \begin{subfigure}{0.45\linewidth}
    \centering
    \includegraphics[width=0.8\linewidth]{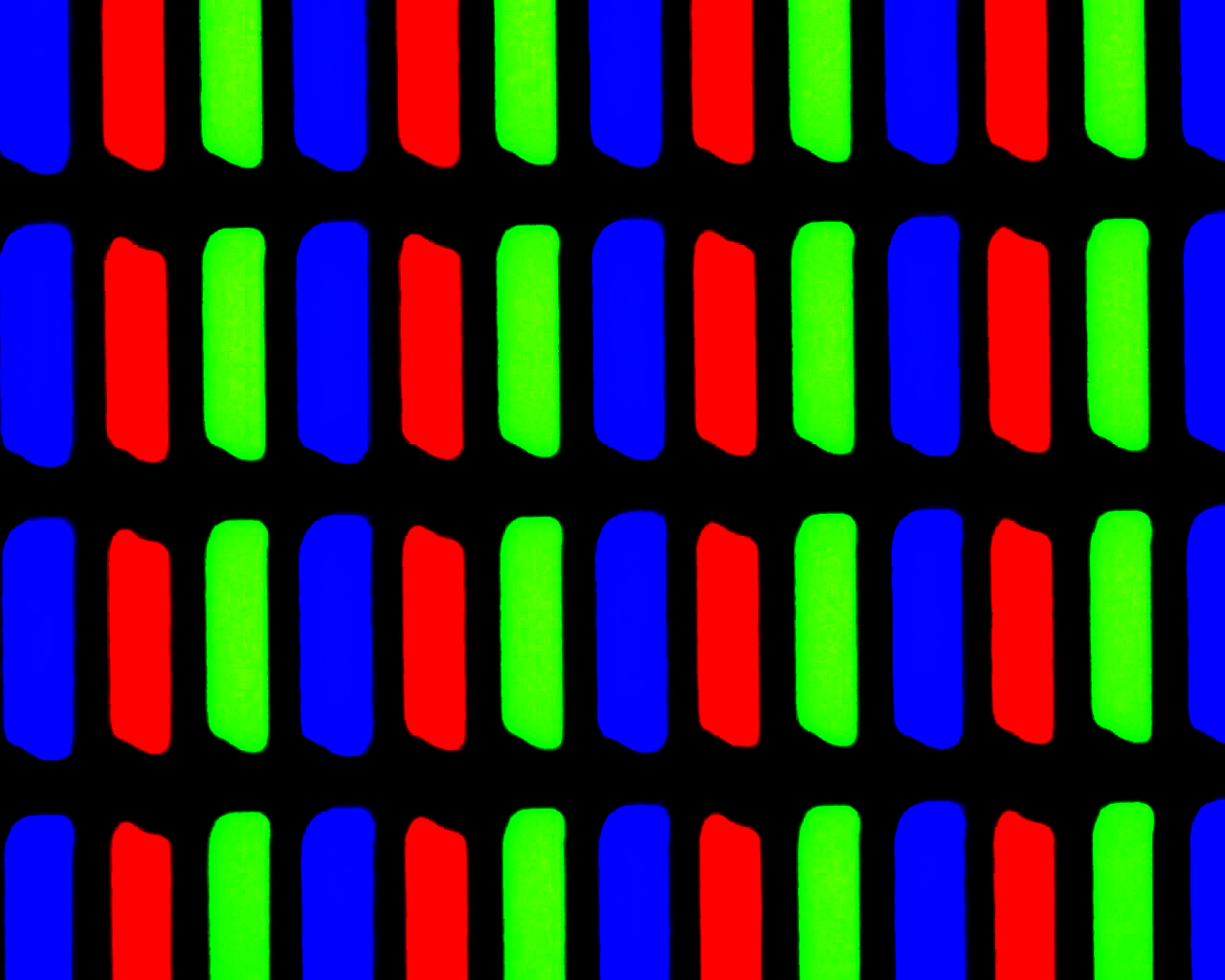}
    \caption{LCD subpixel structure}
    \label{fig:synth_lcd}
  \end{subfigure}
  \begin{subfigure}{0.45\linewidth}
    \centering
    \includegraphics[width=0.8\linewidth]{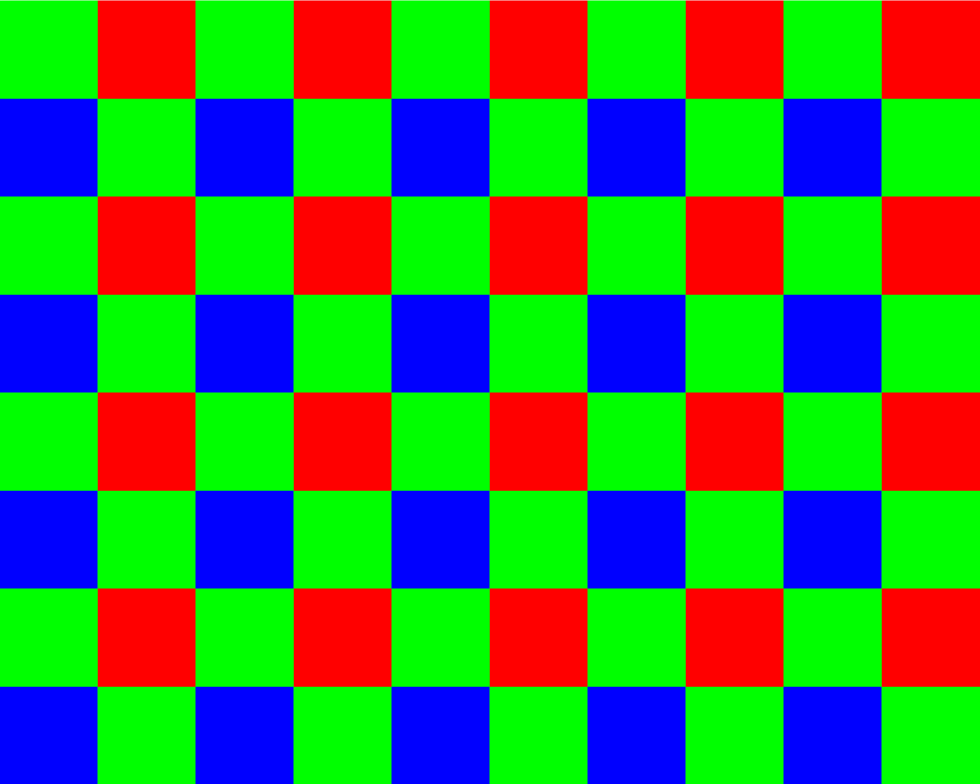}
    \caption{Camera Bayer CFA}
    \label{fig:synth_cam}
  \end{subfigure}
  \caption{Image displayed on a LCD or captured by a camera is
    spatially resampled in order to achieve color effects.}
  \label{fig:synth}
\end{figure}

Considering the drawbacks of using real photos, we employ synthetic
screenshot images with realistic moir\'e patterns for training
instead.  The input images of the synthesizer are collected by using
the print-screen command from computers running Microsoft Windows.  To
accurately simulate the formation of moir\'e patterns, we follow
truthfully the process of image display on an LCD and the pipeline of
optical image capture and digital processing on a camera.

The corresponding ground truth clean image is generated from the
original image using the same projective transformation and lens
distortion function as the moir\'e simulator.  Additionally, the
ground truth image is also scaled to match the same size of the
synthetic camera-captured screen image.

\subsection{Training Details}

From 1,000 digital screenshot images, we create 60,000 $256\times 256$
pairs of artificial moir\'e affected patch and its cleaning ground
truth using the above data synthesizer.  Same as the previous
experiment, of the 60,000 patches 40,000 are used for training, 10,000
are for validation; and another 10,000 are for testing.  For training
the proposed DM-Net, we use $10^5$ update iterations and a batch size
of 32, and we also employ Adam's optimization method
\cite{kinga2015method} with $\beta_1 = 0.9$ and a learning rate of
$10^{-4}$.  To deal with the multiple scales, the loss function of
DM-Net is set as,
\begin{align}
  L_{\ell_2} = \frac{1}{wh}\|g_{0}(y_s) - x\|_2^2 + \frac{1}{WH}\|G_{0}(Y_s) - X\|_2^2,
\end{align}
where the $x$ is the downsampled version of $X$, and $w,h$ are the
corresponding width and height.  The training of DnCNN and RED-Net
follows the settings recommended by the original authors.

Using the proposed two-stage training approach, we retrain DM-Net with
real moir\'e images to get a improved version, namely DM-Net+.  The
training input images for DM-Net+ are real screen images captured
using 3 different smartphones (iPhone 6, iPhone 8, Samsung Galaxy S8)
from various distances and angles.  From the total 300 captured
images, 10,000 $256\times 256$ patches are extracted and used as the
real degraded input $Y$ in the second training stage.  Similar to the
previous super-resolution experiment, the network is trained with
$10^4$ update iterations at a learning rate of $10^{-5}$ in this
stage.  The time required for the second training stage is much less
than the time for the first stage.

For DnCNN and RED-Net, we retrain them using the two-stage approach
with the same configuration as DM-Net+.  As an attempt to further
improve their performances, we extend our training approach to more
stages.  In every stage, the surrogate ground truth is regenerated
using the model from previous stage.  Effectively, this method could
push the results closer to the distribution of the sample clean images
after each stage, due to the minimization of adversarial loss.  But it
can also deviate the results from the degradation model, making them
appear dissimilar to the original images.


\subsection{Experimental Results}
\label{sec:exp_dm}

\begin{figure}[t]
  \captionsetup[subfigure]{labelformat=empty}
  \centering
  \begin{subfigure}{0.24\linewidth}
    \centering
    \includegraphics[width=\textwidth]{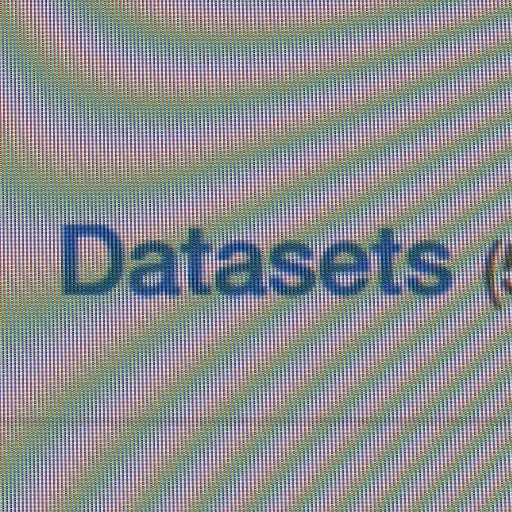}
    \includegraphics[width=\textwidth]{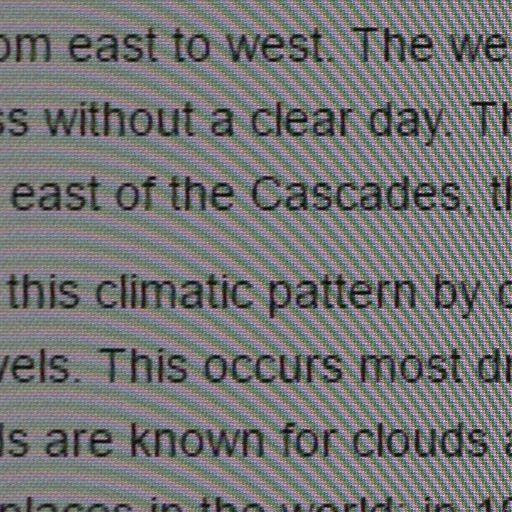}
    \caption{Input}
  \end{subfigure}
  \begin{subfigure}{0.24\linewidth}
    \centering
    \includegraphics[width=\textwidth]{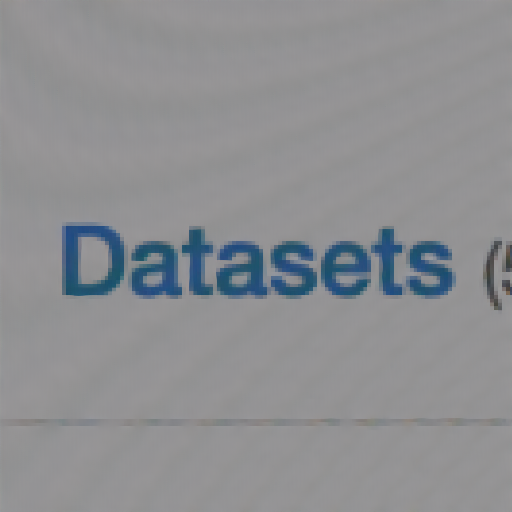}
    \includegraphics[width=\textwidth]{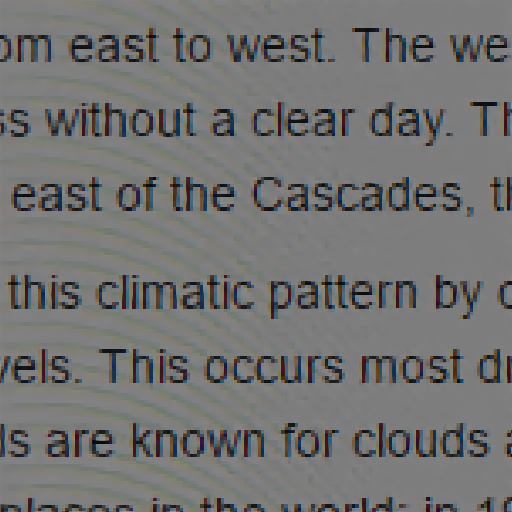}
    \caption{RED-Net}
  \end{subfigure}
  \begin{subfigure}{0.24\linewidth}
    \centering
    \includegraphics[width=\textwidth]{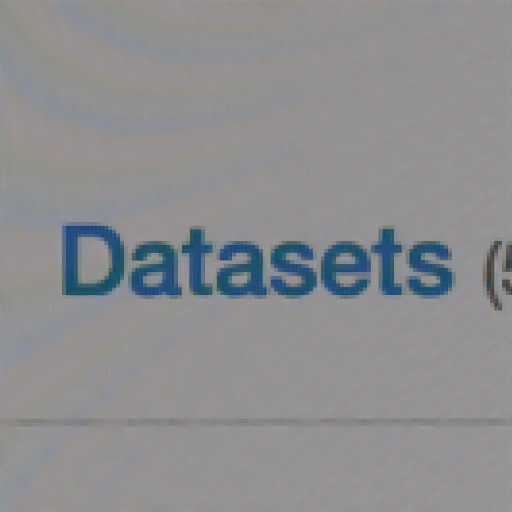}
    \includegraphics[width=\textwidth]{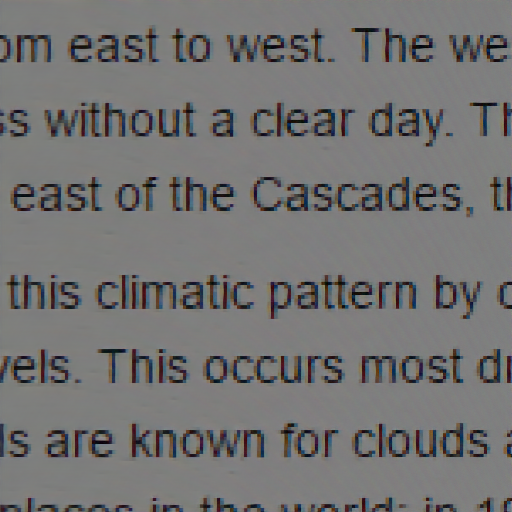}
    \caption{DnCNN}
  \end{subfigure}
  \begin{subfigure}{0.24\linewidth}
    \centering
    \includegraphics[width=\textwidth]{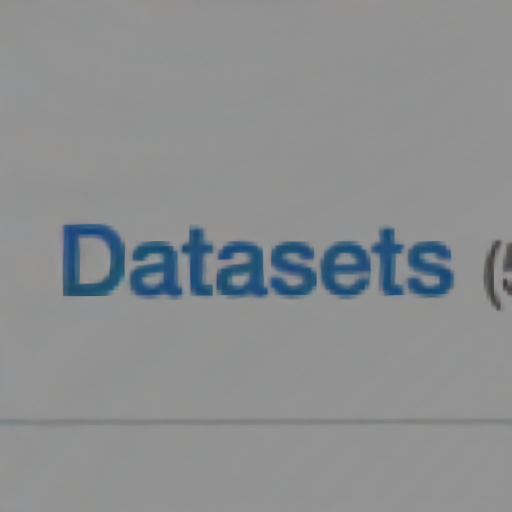}
    \includegraphics[width=\textwidth]{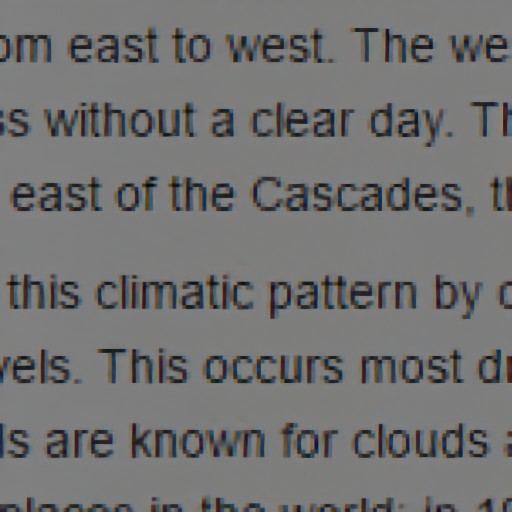}
    \caption{DM-Net}
  \end{subfigure}
  \caption{Results of demoir\'eing techniques for synthetic moir\'e
    images.}
  \label{fig:dm_syn}
\end{figure}


Shown in Fig.~\ref{fig:dm_syn} is the results of the tested techniques
for synthetic input.  The average PSNRs for DnCNN, RED-Net and DM-Net
are 37.46 dB, 37.80 dB and 40.01 dB, respectively.  Both DnCNN and
RED-Net fail to remove the artifacts completely, leaving traces of
wide color bands.  In comparison, the output of DM-Net looks much
cleaner.  As demonstrated by these results, the demoir\'eing problem
is quite challenging; only purposefully designed technique can solve
it successfully.

\begin{figure*}[h]
  \captionsetup[subfigure]{labelformat=empty}
  \centering
  \begin{subfigure}{0.133\linewidth}
    \centering
    \includegraphics[width=\textwidth]{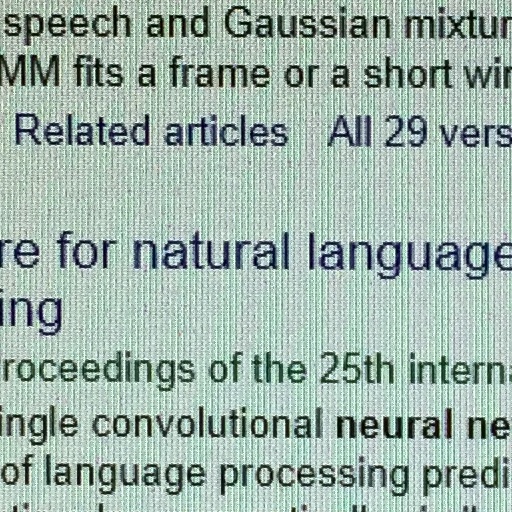}
    \includegraphics[width=\textwidth]{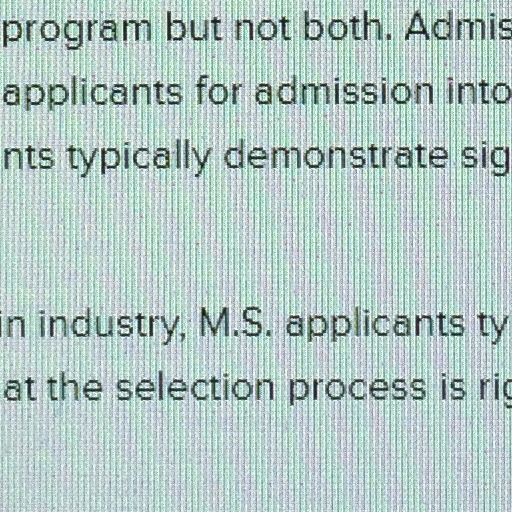}
    \includegraphics[width=\textwidth]{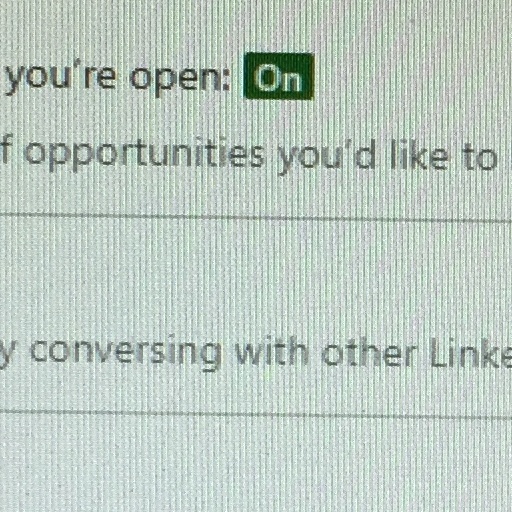}
    \includegraphics[width=\textwidth]{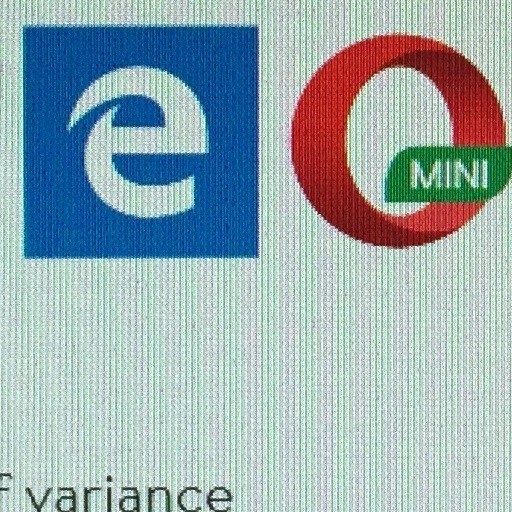}
    \includegraphics[width=\textwidth]{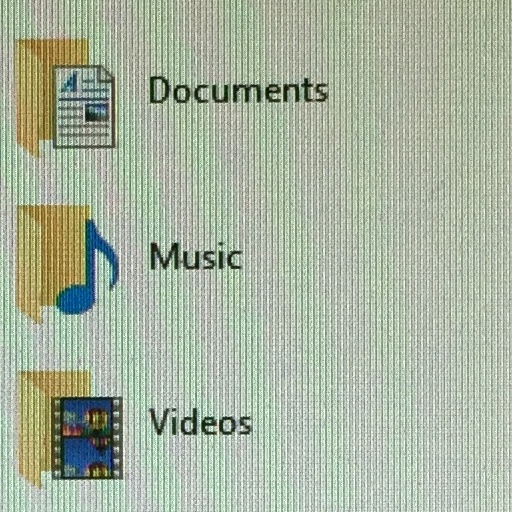}
    \caption{Input}
  \end{subfigure}
  \begin{subfigure}{0.133\linewidth}
    \centering
    \includegraphics[width=\textwidth]{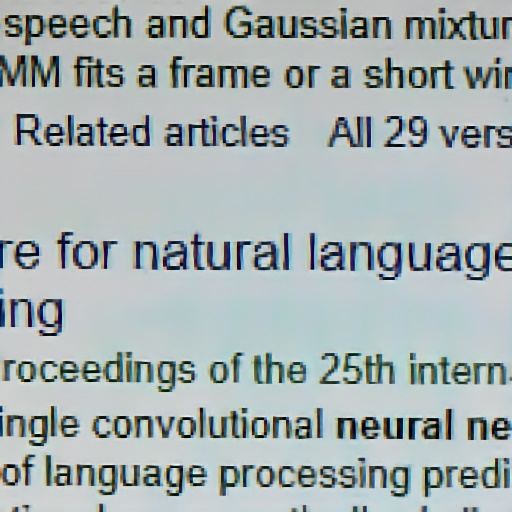}
    \includegraphics[width=\textwidth]{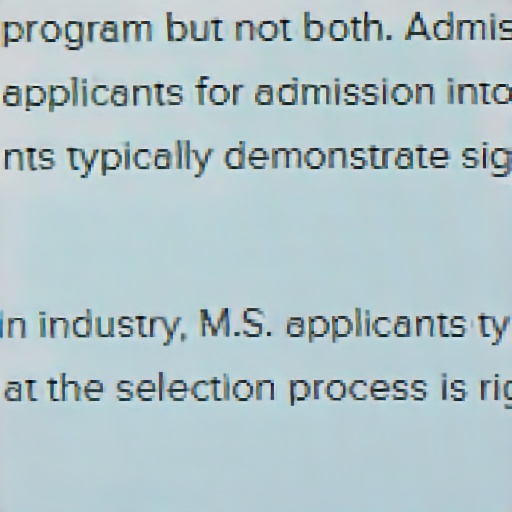}
    \includegraphics[width=\textwidth]{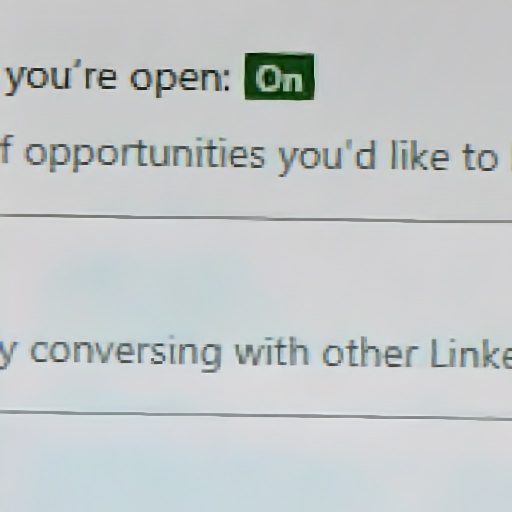}
    \includegraphics[width=\textwidth]{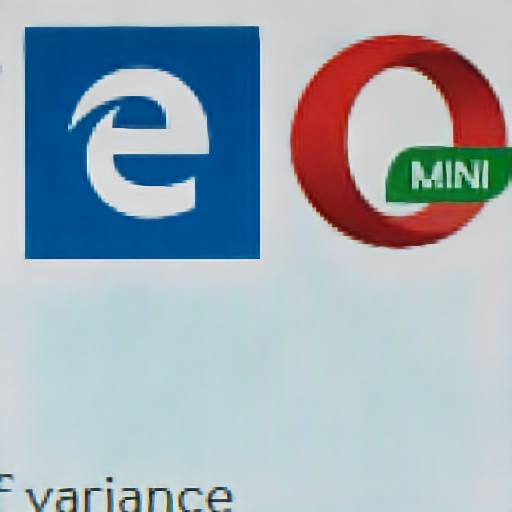}
    \includegraphics[width=\textwidth]{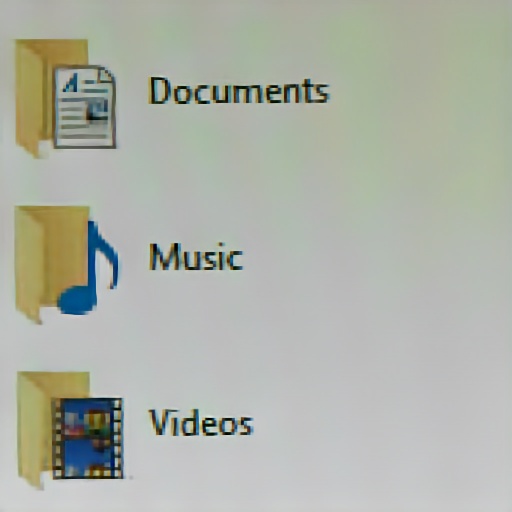}
    \caption{DM-Net}
  \end{subfigure}
  \begin{subfigure}{0.133\linewidth}
    \centering
    \includegraphics[width=\textwidth]{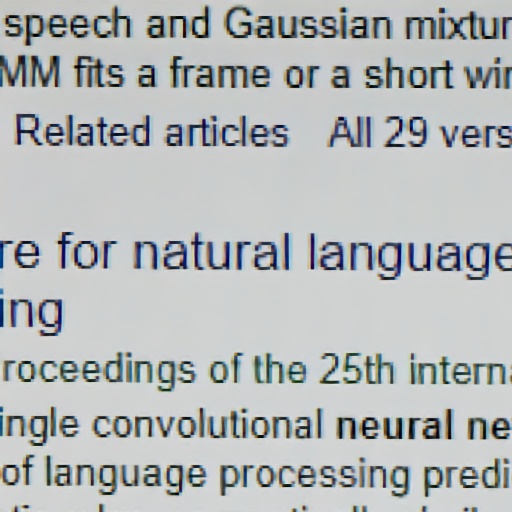}
    \includegraphics[width=\textwidth]{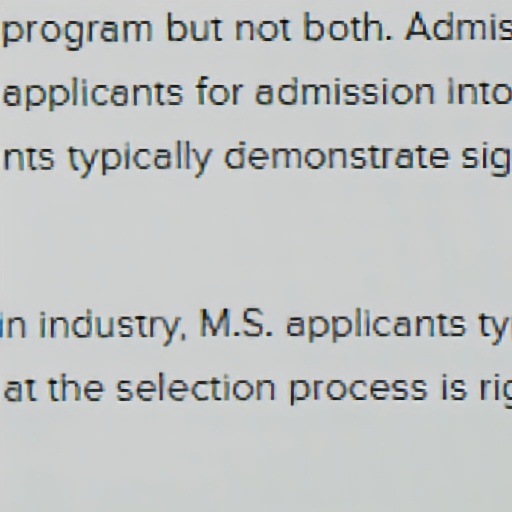}
    \includegraphics[width=\textwidth]{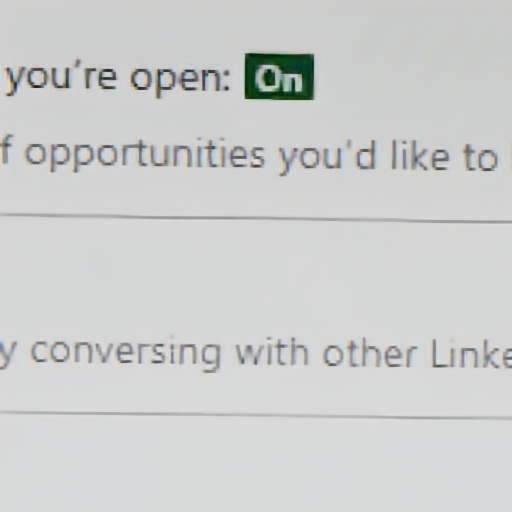}
    \includegraphics[width=\textwidth]{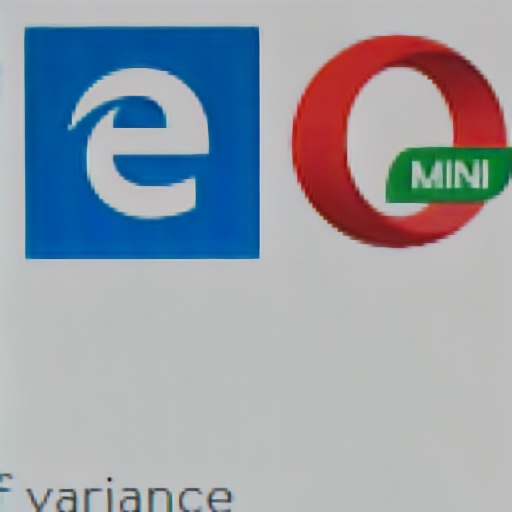}
    \includegraphics[width=\textwidth]{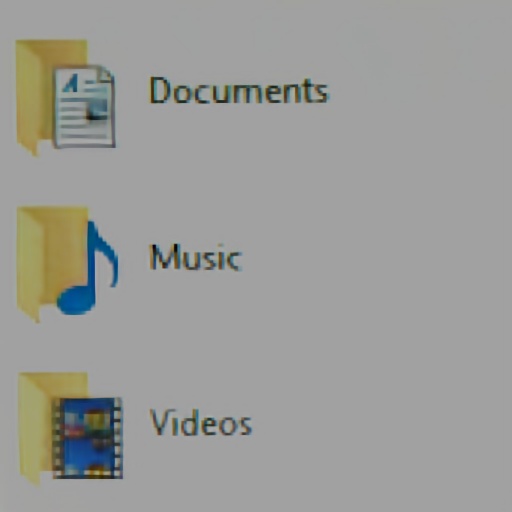}
    \caption{\textbf{DM-Net+}}
  \end{subfigure}
  \begin{subfigure}{0.133\linewidth}
    \centering
    \includegraphics[width=\textwidth]{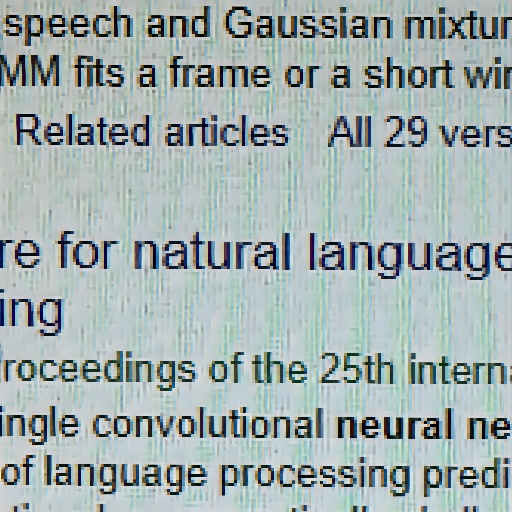}
    \includegraphics[width=\textwidth]{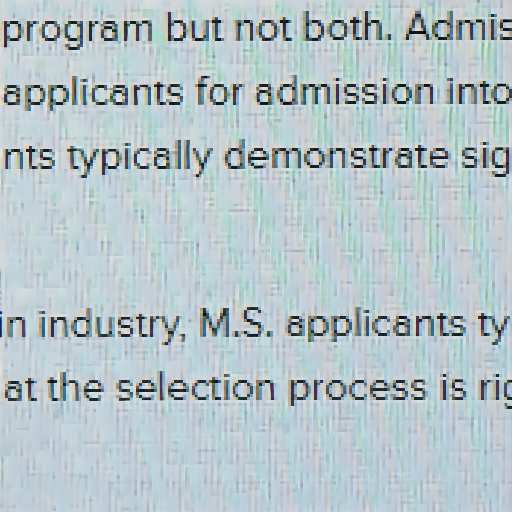}
    \includegraphics[width=\textwidth]{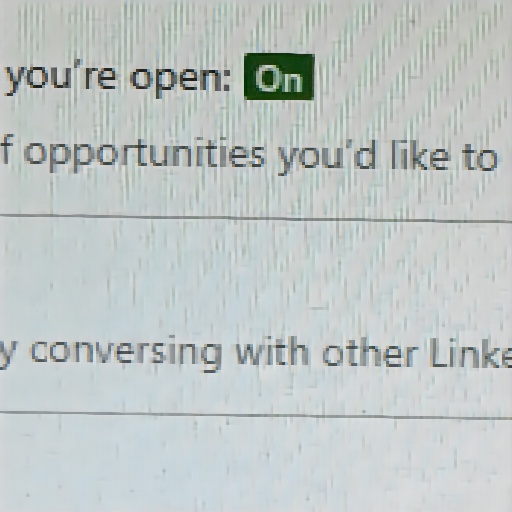}
    \includegraphics[width=\textwidth]{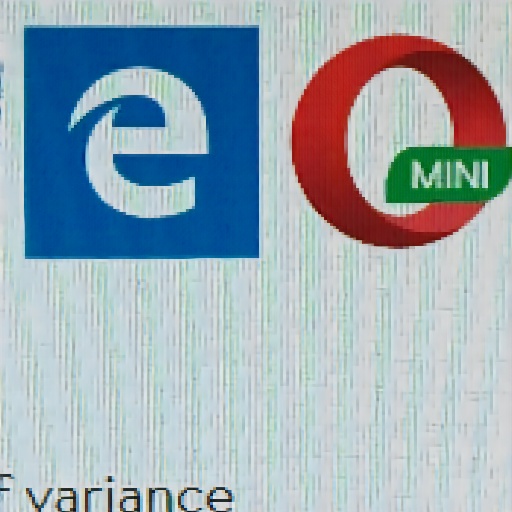}
    \includegraphics[width=\textwidth]{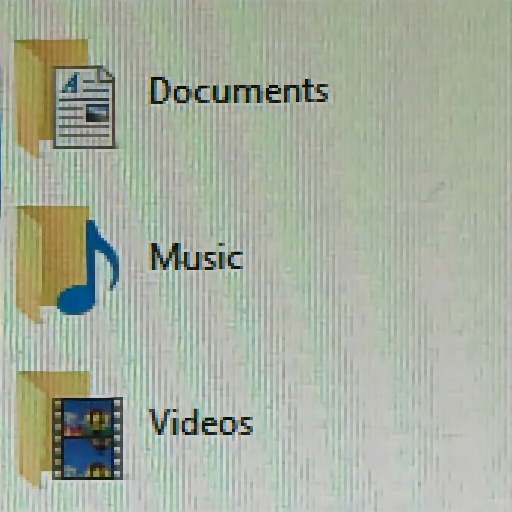}
    \caption{RED-Net}
  \end{subfigure}
  \begin{subfigure}{0.133\linewidth}
    \centering
    \includegraphics[width=\textwidth]{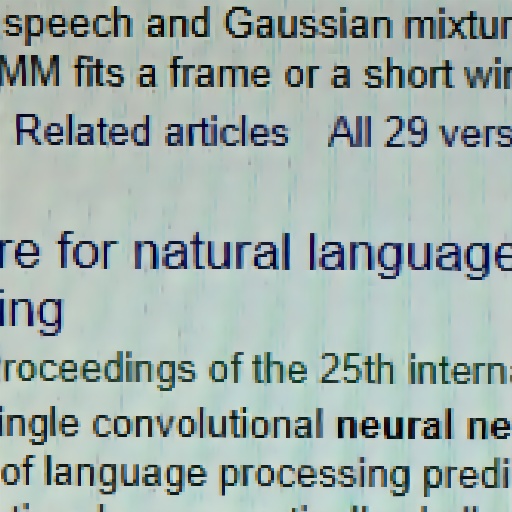}
    \includegraphics[width=\textwidth]{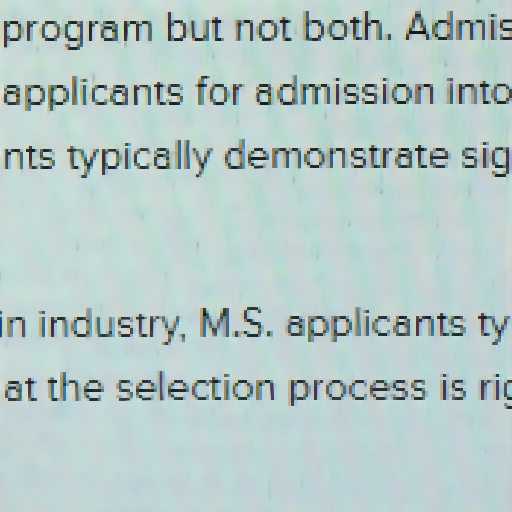}
    \includegraphics[width=\textwidth]{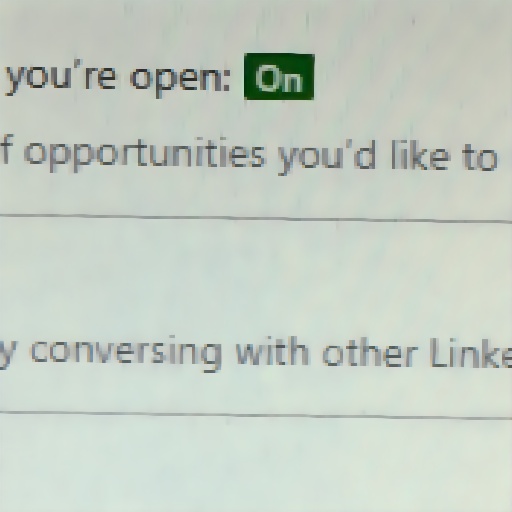}
    \includegraphics[width=\textwidth]{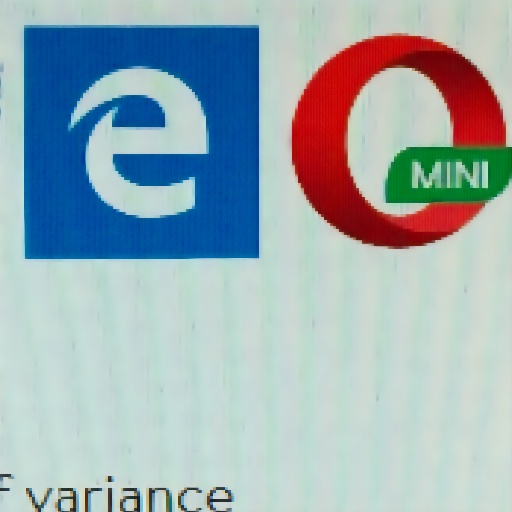}
    \includegraphics[width=\textwidth]{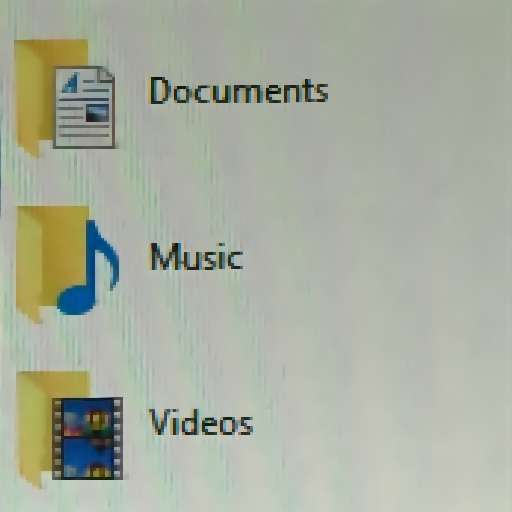}
    \caption{\textbf{RED-Net+}}
  \end{subfigure}
  \begin{subfigure}{0.133\linewidth}
    \centering
    \includegraphics[width=\textwidth]{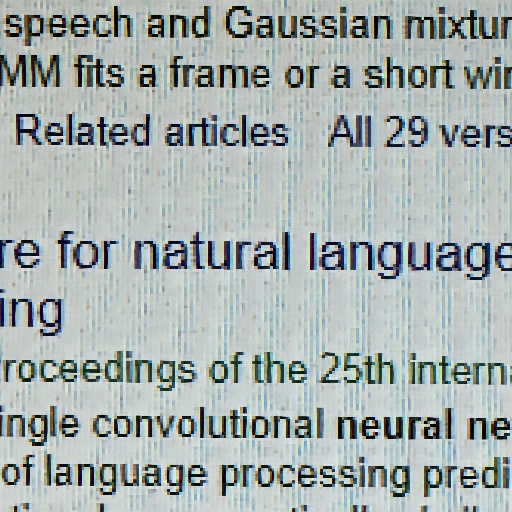}
    \includegraphics[width=\textwidth]{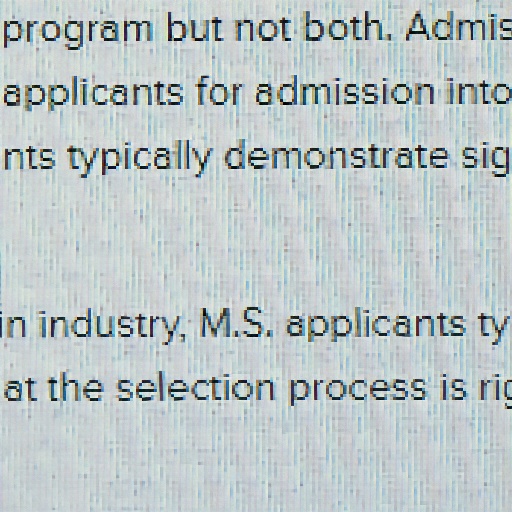}
    \includegraphics[width=\textwidth]{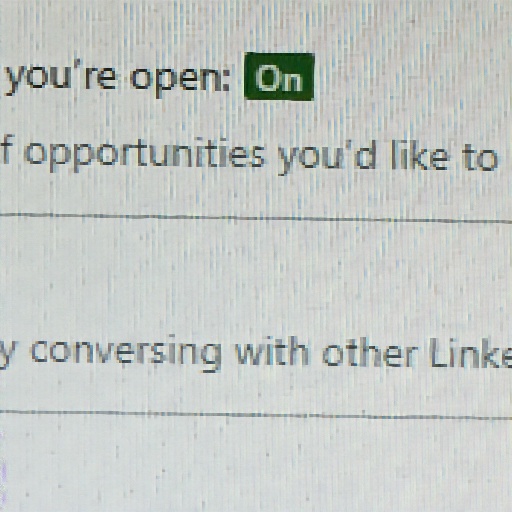}
    \includegraphics[width=\textwidth]{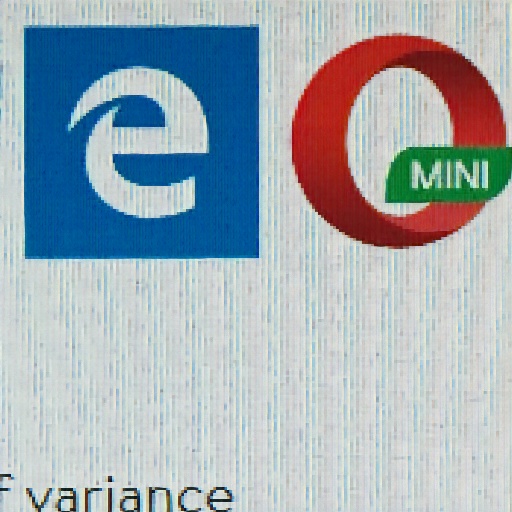}
    \includegraphics[width=\textwidth]{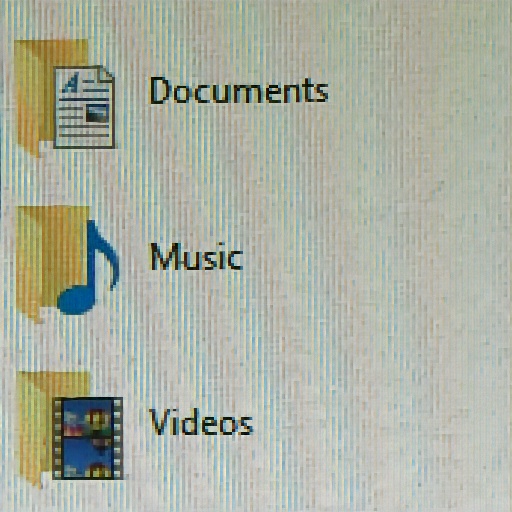}
    \caption{DnCNN}
  \end{subfigure}
  \begin{subfigure}{0.133\linewidth}
    \centering
    \includegraphics[width=\textwidth]{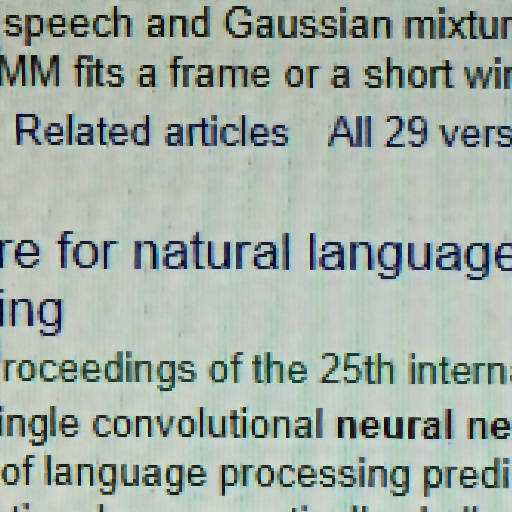}
    \includegraphics[width=\textwidth]{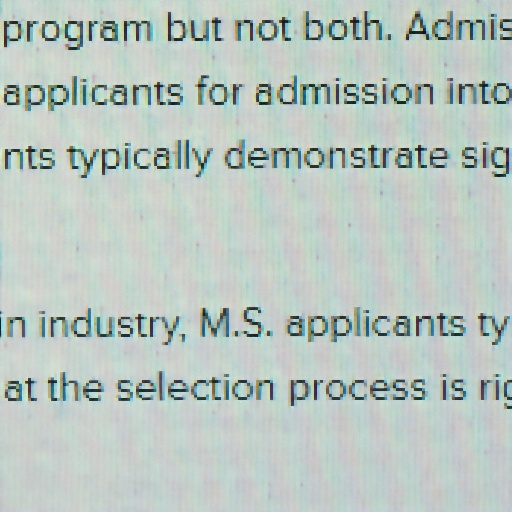}
    \includegraphics[width=\textwidth]{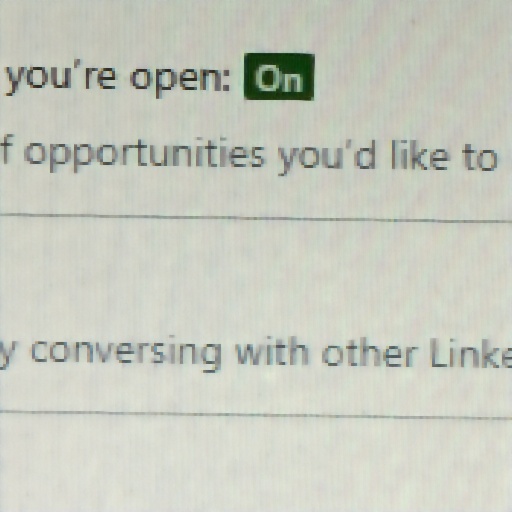}
    \includegraphics[width=\textwidth]{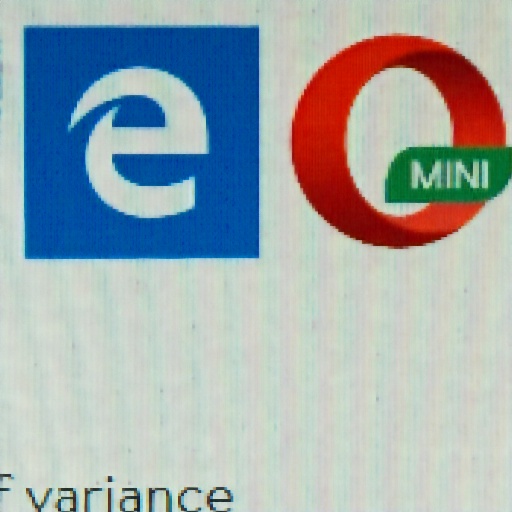}
    \includegraphics[width=\textwidth]{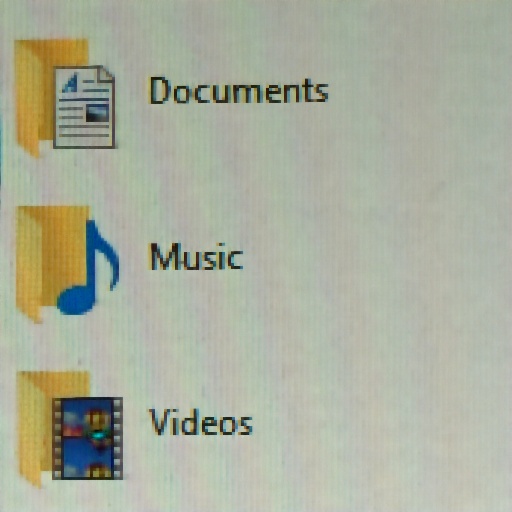}
    \caption{\textbf{DnCNN+}}
  \end{subfigure}
  \caption{Results of demoir\'eing techniques for real moir\'e
    images.}
  \label{fig:comp_g1g2}
\end{figure*}

\begin{figure}[h!]
  \captionsetup[subfigure]{labelformat=empty}
  \centering
  \begin{subfigure}{0.28\linewidth}
    \centering
    \includegraphics[width=\textwidth]{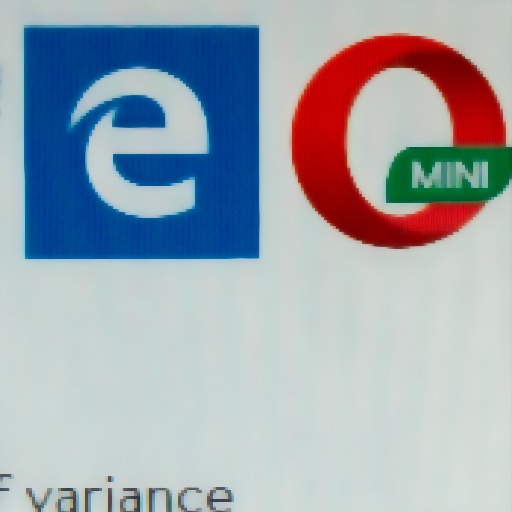}
    \caption{\textbf{RED-Net++}}
  \end{subfigure}
  \begin{subfigure}{0.28\linewidth}
    \centering
    \includegraphics[width=\textwidth]{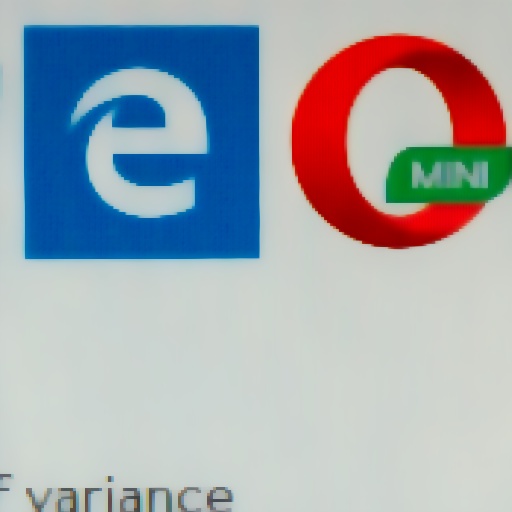}
    \caption{\textbf{RED-Net+++}}
  \end{subfigure}
  \begin{subfigure}{0.28\linewidth}
    \centering
    \includegraphics[width=\textwidth]{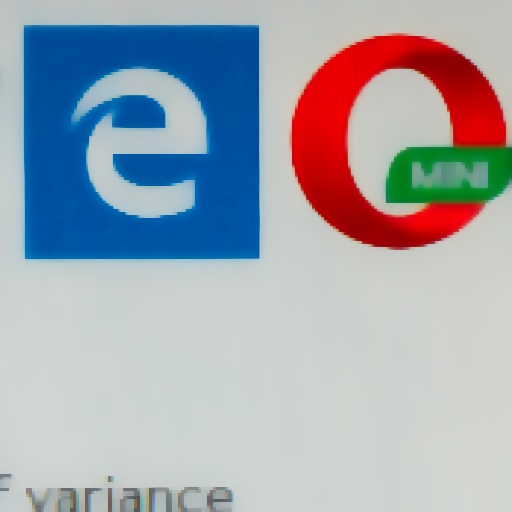}
    \caption{\textbf{RED-Net++++}}
  \end{subfigure}
  \begin{subfigure}{0.28\linewidth}
    \centering
    \includegraphics[width=\textwidth]{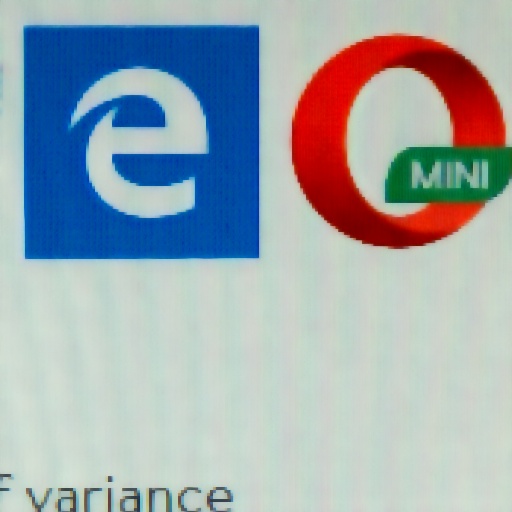}
    \caption{\textbf{DnCNN++}}
  \end{subfigure}
  \begin{subfigure}{0.28\linewidth}
    \centering
    \includegraphics[width=\textwidth]{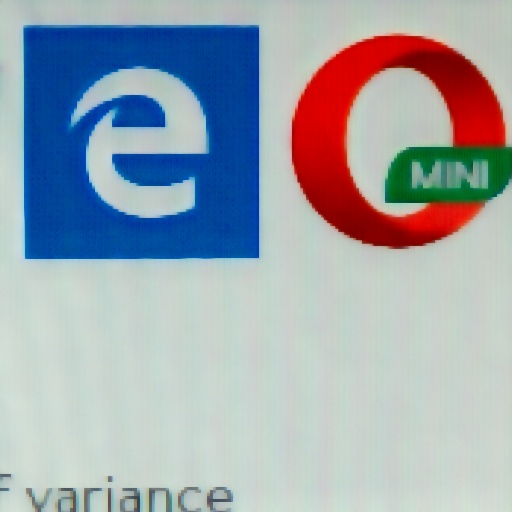}
    \caption{\textbf{DnCNN+++}}
  \end{subfigure}
  \begin{subfigure}{0.28\linewidth}
    \centering
    \includegraphics[width=\textwidth]{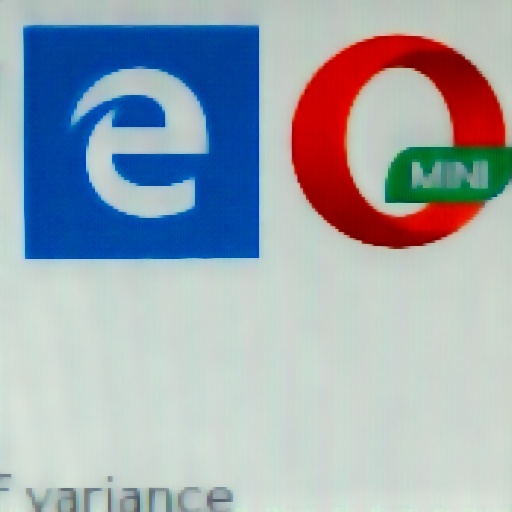}
    \caption{\textbf{DnCNN++++}}
  \end{subfigure}
  \caption{Results of RED-Net and DnCNN using 3-stage, 4-stage and
    5-stage training.}
  \label{fig:comp_plus}
\end{figure}

However, as shown in Figure~\ref{fig:comp_g1g2}, DM-Net, which
achieves excellent performance on the synthetic data, fails to obtain
the same quality results for real camera-captured screen images.  In
contrast, the improved network DM-NET+ works much better for real
input images, leaving almost no artifacts.  The refined models
RED-Net+ and DnCNN+ also make some significant improvements over the
original versions without any change to their architectures.  However,
their results for real images are still relatively noisy in comparison
with the results of DM-NET+.  After all, the architectures of the two
networks are not designed specifically for the demoir\'e task.  We can
further improve RED-Net+ and DnCNN+, to some extent, using more
training stages as discussed previously.  Shown in
Figure~\ref{fig:comp_plus} are some of the results.  With more
training stages, the results become cleaner but also blurrier in
general.



\section{Conclusion}

In many applications of deep learning, we have to resort to
synthesized paired data to train the DCNN, due to the lack of real
data.  However, the most likely scenario for a deep learning
restoration system is that the pairs of the latent and degraded images
can only be generated to a limited precision.  These paired images,
although being flawed, are still valuable initial training data to lay
a basis for supervised deep learning, as in current practice.  In this
work, we are not contented as others with the expediency of artificial
training data, and instead strive to correct the deficiency of the
existing methods by assuming an overly simplified degradation process.
Our main contribution is a more general and robust neural network
training approach that can withstand the discrepancies between the
synthesized degraded images for training and the actual input images.

{\small
\bibliographystyle{ieee}
\bibliography{restoration}
}

\end{document}